\documentclass[10pt]{article}
\usepackage[preprint]{tmlr}
\usepackage{booktabs}
\usepackage{amsmath}
\usepackage{amssymb}
\usepackage{graphicx}
\usepackage{subcaption}
\usepackage{tabularx}
\usepackage{array}
\usepackage{placeins}
\usepackage{flafter}
\usepackage{hyperref}
\usepackage{url}

\title{Pixel Decodability Is Not a Compression Signal: Causally Evaluating Importance Proxies for Visual KV-Cache Eviction}
\author{\name Chenyu Zhou \email zhou.c.76d6@m.isct.ac.jp \\
\addr School of Engineering, Institute of Science Tokyo, Japan
\AND
\name Qiliang Jiang \email jiangqiliang@zju.edu.cn \\
\addr College of Control Science and Engineering, Zhejiang University, China
\AND
\name Shuning Wu \email shuningwu@u.nus.edu \\
\addr Department of Electrical and Computer Engineering, National University of Singapore
\AND
\name Xu Zhou \email zhouxu\_nus@u.nus.edu \\
\addr Department of Electrical and Computer Engineering, National University of Singapore}

\def\month{MM}
\def\year{YYYY}
\def\openreview{\url{https://openreview.net/forum?id=XXXX}}

\begin{document}
\maketitle

\begin{abstract}

Vision--language models (VLMs) retain, in the visual key--value (KV)
cache of their language model, a substantial amount of pixel-decodable
visual content. Yet we show, in our setting, that this retention is
\emph{task-inert}: across our preregistered tests, how much decodable
content a unit retains never positively tracks whether the computation
that answers the question causally relies on that unit --- where
retention relates to task structure at all, the association is weak and
points the wrong way. We quantify how much decodable content
each visual unit \emph{retains} with a learned pixel-inversion decoder,
and how much the model \emph{causally uses} it with single-super-patch
KV ablation (the teacher-forced drop in gold-answer log-probability),
and relate the two within images under a preregistered, sign-calibrated,
held-out design. Retention is decoupled from attention
(\(\rho\approx-0.07\) to \(-0.09\)) and, in a preregistered, well-powered
null (held-out \(N{=}72\), power \(\approx0.9\); upper CIs exclude
\(\rho\gtrsim0.06\)/\(0.01\)), from causal utilization (\(\rho\approx-0.01\) to
\(-0.05\)). Yet utilization is not
inert to every proxy: attention weakly but significantly tracks it
(\(\rho\approx+0.11\) to \(+0.12\), held-out intervals excluding zero)
--- the only signal we find that does, serving as the design's
positive control. We characterize pixel-decodable retention as an
\emph{informational axis} of the visual KV cache, orthogonal to the
functional (attention--utilization) one. How much of this task-inert content a
cache holds differs by architecture in our model pair: the encoder-free
VLM retains \(2.68\times\) more than the encoder-based one. The
engineering consequence is a controlled negative result: at super-patch
granularity, deconfounded pixel-decodable retention ranks KV eviction no
better than random; at token granularity it acquires only a weak
inverse-importance signal at the larger budgets --- dominated at every
budget by attention magnitude, the weak-but-real proxy.
In our setting, pixel-decodable reconstructability is not a competitive
KV-compression signal at any granularity we test.

\end{abstract}

\section{Introduction}\label{sec:1}

Vision--language models (VLMs) carry a large visual key--value (KV)
cache (hundreds of tokens per image, across every layer), and a
growing body of work compresses it by \emph{evicting} the visual units
deemed least important. The importance signal is almost always a proxy:
the magnitude of attention a unit receives, or the reconstructability of
its representation. Both encode an implicit assumption: that the
proxy tracks what the model actually \emph{uses}. We ask whether that
assumption holds by taking the assumption apart into three distinct
questions about a visual unit's KV. How much decodable pixel content
does it \emph{retain}? How much attention does it \emph{receive}? And
how much does the model \emph{causally rely} on it to produce the
correct answer? These are routinely conflated; we measure them
separately.

\begin{figure}[t]
  \centering
  \includegraphics[width=\linewidth]{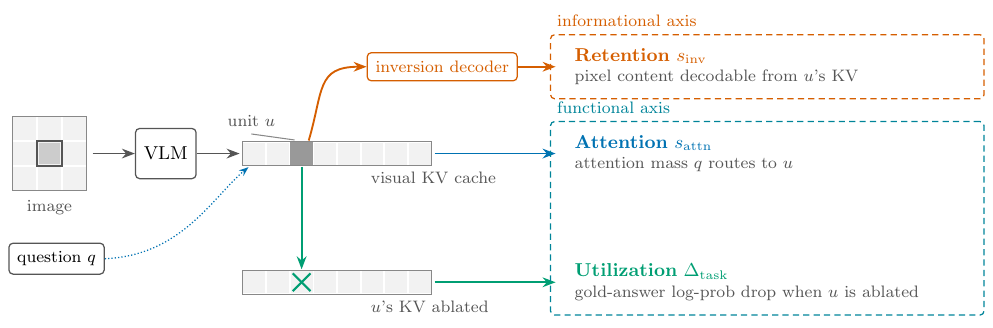}
  \caption{Three readouts of one visual KV unit $u$. \emph{Retention} ($s_\mathrm{inv}$): a learned pixel-inversion decoder measures how much of $u$'s image content its KV still decodes to pixels --- the informational axis. \emph{Attention} ($s_\mathrm{attn}$): the attention mass the question routes to $u$. \emph{Utilization} ($\Delta_\mathrm{task}$): the drop in gold-answer log-probability when $u$'s KV is ablated --- together with attention, the functional axis. The paper measures the relationships between the two axes within images; estimates are in Figure~\ref{fig:forest}.}
  \label{fig:schematic}
\end{figure}

We instrument two of these axes directly. \emph{Retention} is read out
by a learned decoder that inverts a unit's KV back to pixels, a
content-level probe, blind to the attention pattern (\S\ref{sec:31}).
\emph{Utilization} is read out causally: we ablate a single
super-patch's KV and measure the teacher-forced drop in the gold
answer's log-probability, a per-unit causal readout of ``does removing
exactly this unit hurt the answer'' (\S\ref{sec:32}). Relating the two within each
image, under a preregistered partial-correlation design with sign
calibration on a held-out split (\S\ref{sec:34}), we find a clean dissociation.
Figure~\ref{fig:schematic} lays out the three readouts
and the relationships we measure between them.

Our findings, on TextVQA with an encoder-free (Gemma-4-12B) and an
encoder-based (InternVL3.5-8B) VLM, are three. (i) Pixel-decodable
retention predicts neither attention (\(\rho\approx-0.07\) to
\(-0.09\)) nor causal utilization: the covariate-adjusted partial
correlation is \(-0.01\)/\(-0.05\) across the two models with CIs
including zero, under a design powered at \(\approx 0.9\) for the
preregistered effect size of \(\rho=0.15\), and the hypothesized
positive coupling never appears. (ii) Attention, by contrast, does carry
a weak but genuine utilization signal (\(+0.11\) to \(+0.12\), held-out
CIs excluding zero), and because the very same estimator detects it,
the retention null reflects retention itself rather than an insensitive
instrument. Retention is orthogonal to this functional signal; we cast
pixel-decodable retention as an \textbf{informational axis} of the
visual KV cache: content the cache carries and that is extractable,
but that the task computation does not route through. (iii) How much of
this task-inert content a cache holds differs by architecture \emph{in
our model pair}: the encoder-free model retains \(2.68\times\) more than
the encoder-based one. In the eviction experiment, the ordering
predicted by (ii) is recovered on the primary causal-cost metric:
attention-guided eviction beats random; deconfounded retention does
not, in either direction of the ranking, on the super-patch grid ---
and at token granularity it helps only through a weak \emph{inverse}
signal (\S\ref{sec:45}).

\textbf{Contributions.} (1) We define a \emph{per-unit causal readout}
of utilization (single-super-patch KV ablation) and use it to
evaluate importance proxies directly, rather than through downstream
accuracy. (2) We jointly quantify retention, attention, and utilization
in the VLM visual cache, turning a two-way intuition into a three-way
analysis that repositions retention as an informational axis orthogonal
to the functional one rather than merely ``decoupled.'' (3) We give a
quantitative model-pair contrast (\(2.68\times\)) associating the
encoder-free/encoder-based distinction with how much task-inert pixel
content the cache retains. (4) We contribute a controlled negative
result: deconfounded pixel-decodable retention ranks KV eviction no
better than random at super-patch granularity, and acquires only a
weak inverse-importance signal at token granularity, providing causal
evidence, in our setting, against the assumption that pixel-level
reconstructability implies importance.

The pixel-inversion instrument,
and the finding that encoder-free VLMs reconstruct more from deep KV
than encoder-based ones, originate in earlier privacy-oriented work
\citep{priorinversion}; the retention/utilization \emph{dissociation},
the causal-utilization readout, the three-axis analysis, and the
compression falsification are new here; the prior work quantified
\emph{what can be reconstructed}, not whether it is \emph{used} (\S\ref{sec:2}).

\section{Related Work}\label{sec:2}

\textbf{Decodability is not causal use.} A recurring finding across
representation analysis is that information \emph{recoverable} from a
hidden state by a trained probe need not be information the network
\emph{uses} to produce its output. Probing studies in vision
transformers recover object counts and spatial attributes that the model
does not causally rely on for its predictions \citep{countingvits}, and
analogous dissociations appear in language models, where structural
information (e.g., bracket-matching state) is linearly decodable yet
causally inert for the behavior it superficially predicts
\citep{bracketseq}. Methodologically, our utilization readout is in the
ablate-and-measure lineage of amnesic probing \citep{amnesicprobing} ---
intervene on a representation and measure the behavioral change ---
applied at the granularity of a single visual KV unit. We inherit this
decodable-\(\neq\)-causal lens and instantiate it for a concrete engineering
object, the visual key--value (KV) cache of a vision--language model
(VLM), where the gap between what is retained and what is used has
direct consequences for cache compression.

\textbf{Retention, accessibility, and utilization in text-only caches.}
Closest in vocabulary is recent work formalizing \emph{retention},
\emph{accessibility}, and \emph{utilization} of information in the KV
cache of text-only LLMs \citep{physicskv}. We deliberately keep our
terminology aligned to make the comparison legible, but the
instrumentation and the object of study differ on every axis. Their
setting is text tokens; ours is visual tokens carrying pixel content.
Their retention is read out with linear concept-vector probes and their
accessibility with attention-derived graph reachability; ours is read
out with an \emph{independent, learned pixel-inversion decoder} --- a
content-level probe that never sees the attention pattern. Their
utilization is inferred from downstream task accuracy; ours is a
\emph{per-unit causal readout} obtained by ablating a single visual
super-patch's KV and measuring the teacher-forced drop in gold-answer
log-probability. The shared vocabulary is a feature: it lets us state
precisely that the retention/utilization dissociation we document is a
\emph{visual, pixel-level, causally-grounded} instance, not a re-run of
the text-only result.

\textbf{Reconstruction as a compression signal for VLM caches.} A
parallel line of work compresses VLM KV/token caches and, in several
cases, uses \emph{reconstructability} of the compressed representation
as the pruning signal: retaining the units whose features can be
recovered \citep{spare}, or regressing value vectors to decide what is
redundant \citep{vector}; in text-only KV caches, reconstruction of the
full context has likewise been used as a query-agnostic eviction score
\citep{kvzip}. These methods reconstruct in \emph{feature}
space (learned subspaces, value regressions, context re-generation) and
report genuine compression payoffs there. Our probe operates
one level lower (at the level of \emph{pixel-decodable content},
using a decoder trained only to invert KV back to the image), and we
use it \emph{diagnostically} (to characterize an axis of the
representation) rather than as a scoring function. The distinction
matters for our negative result: it is specific to pixel-decodable
content, while feature-space reconstruction targets a different, and
evidently more task-aligned, quantity. Attention-magnitude pruning --- FastV-style eviction of low-attention
visual tokens \citep{fastv} and its cache-eviction variants
\citep{vlcache,retentivekv}, where the latter's ``retention'' denotes a
cache-persistence policy distinct from the pixel-decodable retention
studied here --- is the standard baseline our compression analysis
compares against.

\textbf{Prior work: KV$\to$pixel inversion.} The pixel-inversion decoder we
use as a measurement instrument was introduced by \citet{priorinversion},
which framed KV-cache reconstructability as a \emph{privacy} boundary
and characterized \emph{what} is reconstructable (spatial grid
resolution, character readability, structural fidelity), showing that
encoder-free VLMs reconstruct substantially more from deep KV than
encoder-based ones. That work is entirely about
reconstruction quality; it does not ask, and does not address, whether
the reconstructable content is \emph{causally used} by the model. While the
inversion instrument and the architecture-differential reconstruction
observation originate in \citet{priorinversion}, what is new here is
the question that work did not pose: (i) a \emph{causal readout} of utilization
(single-unit KV ablation) and the finding that reconstructable retention
does not predict it --- a preregistered, well-powered dissociation, not
a qualitative aside; (ii) the \emph{attention} axis, turning a two-way
measurement into a three-way analysis that positions retention as an
informational axis orthogonal to the functional one (\S\ref{sec:43}); and (iii)
the reframe from a privacy attack into a cache-compression diagnostic,
with a controlled eviction falsification (\S\ref{sec:45}). The instrument is prior
art; the dissociation, its causal grounding, and its compression
consequence are not.

\textbf{Limits of attention-based visual pruning evaluations.} Finally,
recent evaluation work shows that attention-guided visual-token pruning
carries spatial biases and can degrade on tasks requiring peripheral
evidence \citep{feather}, complementing our finding that attention is
only a \emph{weak} predictor of causal utilization (\S\ref{sec:43}): it is the
better of the two proxies we study, but it is not a strong one.

\section{Method}\label{sec:3}

We measure three axes of a VLM's visual KV cache: how much pixel
content it \emph{retains}, how much attention it \emph{receives}, and
how much the model \emph{causally uses} it --- and relate them within a
preregistered partial-correlation design.
All quantities are defined per \emph{super-patch} (a spatial unit of the
visual grid) and estimated within-image to remove image-level confounds.

\subsection{Retention: learned pixel
inversion}\label{sec:31}

For a fixed decoder layer \(\ell\), we read the visual KV of an
image-only prefill (the image with an empty/neutral prompt, so that KV
reflects encoded visual content rather than question-conditioned state)
and train an inversion decoder to reconstruct the input image from that
KV. The retention score \(s_\mathrm{inv}\) of a super-patch is the
reconstruction quality attributable to that patch's KV, measured as a
\emph{shuffle-adjusted excess}: the reconstruction score with the
patch's KV in place minus the score under a within-image shuffle of KV
positions, so that \(s_\mathrm{inv}\) isolates position-specific
decodable content from what any KV would afford on average. We report
the main analysis at the primary layer \(\ell = 0\) (the layer at
which the instrument is trained and validated); deeper layers are
secondary. To quantify measurement noise, we train the decoder under
\emph{two independent seeds} and take the two resulting
\(s_\mathrm{inv}\) maps as parallel measurements of the same latent
quantity (used for disattenuation, \S\ref{sec:34}). Decoder
architecture, training protocol, and the shuffle-adjusted score are
specified in Appendix~\ref{app:instrument}.

\subsection{Utilization: single-unit causal
ablation}\label{sec:32}

Retention asks what can be \emph{read out} of the KV; utilization asks
what the model \emph{would lose} without it. For each super-patch we
ablate its visual KV (masking those key/value positions so downstream
attention cannot attend to them) and measure the change in the
teacher-forced gold-answer log-probability:
\[\Delta_\mathrm{task}(u) \;=\; \log p(\text{gold} \mid \text{full KV}) \;-\; \log p(\text{gold} \mid \text{KV} \setminus u),\]
summed over the gold answer's tokens. A large positive
\(\Delta_\mathrm{task}\) means the patch's KV was causally load-bearing
for the correct answer; \(\Delta_\mathrm{task}\approx 0\) means it was
inert. This is a \emph{per-unit causal readout} of utilization, as
opposed to accuracy-based proxies: it directly interrogates the
counterfactual ``does removing exactly this unit hurt the gold answer.''
That \(\Delta_\mathrm{task}\) carries genuine signal rather than noise
is confirmed downstream (\S\ref{sec:43}): attention weakly but significantly
predicts it in both models (held-out cluster-bootstrap CIs exclude
zero). This serves as the design's positive control (\S\ref{sec:42});
an external criterion validation --- $\Delta_\mathrm{task}$ localizes
OCR-grounded answer regions on both models --- is in
Appendix~\ref{app:criterion}. The readout is a
\emph{single-unit} intervention, so it can understate importance that is
redundantly distributed across units, particularly after the ViT's
bidirectional mixing in the encoder-based model (whose per-unit
\(\Delta_\mathrm{task}\) is therefore a lower bound). We pair it with
cumulative eviction (\S\ref{sec:45}) and with a whole-image ablation
arm, which removes \emph{all} visual KV and yields mean gold-logprob
drops of $8.1$/$9.9$ for Gemma/InternVL (positive on $94\%$/$90\%$ of
images): the small per-patch effects reflect distributed reliance.

\subsection{Attention readout}\label{sec:sattn}

The third axis, attention \(s_\mathrm{attn}(u)\), is the mean attention
mass a super-patch's KV positions receive from the question's text
tokens in the question-conditioned forward pass: for each of the model's
full-attention layers (5, 11, and 17 in both models), we average the
attention weights over heads and over the question's token positions to
obtain a per-visual-token mass, take the equal-weight mean over the
three layers, and sum over the tokens assigned to the super-patch. This
is the FastV-style text\(\to\)vision importance signal
\citep{fastv} that attention-based eviction methods rank by, computed at
the same granularity as the other two readouts.

\subsection{Unit of analysis and
models}\label{sec:33}

We study two VLMs chosen to span the architectural axis of interest:
\textbf{Gemma-4-12B}, an \emph{encoder-free} VLM (pixels enter the
language model directly), whose visual grid varies with aspect ratio
(e.g., \(13\times 20\); 252--275 tokens per image over our
images); and \textbf{InternVL3.5-8B}, an
\emph{encoder-based} VLM (a ViT encoder mediates the visual input),
whose single tile yields \(16\times16 = 256\) tokens. Both grids are
aggregated to a \(4\times4 = 16\) super-patch grid per image; because
Gemma's grid does not divide evenly into \(4\times4\), its
token-to-super-patch map is \emph{non-uniform} (block sizes such as
\([20,20,20,20,15,\dots]\) for a \(13\times20\) image), and we use the
exact block assignment throughout rather than assuming equal-sized
patches. Super-patch
aggregation trades spatial resolution for statistical power: a per-token
unit would leave too few independent units per image for a within-image
correlation, whereas 16 super-patches give a usable within-image sample
while still localizing to regions; the coarser cumulative-eviction
analysis (\S\ref{sec:45}) provides a complementary check at the group level.
Evaluation is on the official TextVQA validation set, a text-reading VQA
task where the answer is localized to specific image regions. This
localization is load-bearing for the instrument: it is what makes
single-unit ablation informative.

\subsection{Estimand: within-image rank partial
correlation}\label{sec:34}

Our primary quantity is the \emph{within-image rank partial correlation}
between retention and utilization. Within each image we rank
super-patches by \(s_\mathrm{inv}\) and by \(\Delta_\mathrm{task}\), and
we residualize both on a fixed covariate set (attention mass
\(s_\mathrm{attn}\), two pixel-energy covariates
(\(s_\mathrm{pixel\text{-}edge}\), \(s_\mathrm{pixel\text{-}entropy}\)),
and spatial position (super-patch row and column,
center distance), so that the reported association is \emph{not}
explained by ``this patch has more texture,'' ``this patch is attended
to,'' or ``this patch is central.'' Rank residualization is done per
image (16 units) and never across images. Uncertainty is quantified by a
\emph{cluster bootstrap over images} (\(n_\mathrm{boot}=10{,}000\)),
which respects the fact that the 16 super-patches of an image are not
independent.

\textbf{Sign calibration and held-out confirmation.} To avoid a
post-hoc sign selection, we fix the \emph{sign} of the hypothesized
effect on a calibration split and \emph{test} it on a disjoint
validation split. Image IDs are partitioned three ways ---
decoder-training / calibration (\(N=48\)) / validation (\(N=72\)) ---
pairwise disjoint, so no image contributes to both the instrument and
the test. The sign \(\mathrm{sign_{cal}}\) is frozen from calibration;
the verdict is read only on validation; the frozen sign is the sign of
the Gemma calibration estimate, shared across both models (\S\ref{sec:42}
reports both models' calibration values).

\textbf{Disattenuation.} Because \(s_\mathrm{inv}\) is measured with
noise, an observed partial correlation understates the true one. Using
the two decoder seeds (\S\ref{sec:31}) we estimate the split-half reliability of
\(s_\mathrm{inv}\) and apply a \emph{conservative} disattenuation ---
dividing only by \(\sqrt{\mathrm{rel}_\mathrm{inv}}\) (correcting for
retention-side noise only), so that the reported effect is, if anything,
an over- rather than under-estimate. This makes a null result robust: a
corrected estimate near zero cannot be dismissed as measurement
attenuation.

\textbf{Power and effect-size bounds.} The design was preregistered with a
smallest effect size of interest of \(\rho = 0.15\) for the
retention--utilization test and a decoupling threshold of \(0.5\) for
C1. The confirmatory
test reads the held-out validation split (\(N=72\); the sign is fixed on
the disjoint 48-image calibration split); at \(N=72\) the
cluster-bootstrap design achieves power \(\approx 0.9\) against that
SESOI. Beyond power, the interval itself bounds the effect: the
two-sided upper limits exclude \(\rho \gtrsim 0.06\) (Gemma) and
\(\rho \gtrsim 0.01\) (InternVL), so we can rule out even modest
positive associations, not merely the SESOI. A null under these
conditions is a \emph{well-powered, bounded} null, not an underpowered
absence of evidence. Two qualifications are stated once here and hold
throughout. Power was computed by plasmode simulation: synthetic effects
of the preregistered size were injected into the empirical per-image
$\Delta_\mathrm{task}$ structure and recovered with the identical
rank-partial, cluster-bootstrap pipeline. And all correlation bounds are
on the \emph{measured} scale: the per-unit $\Delta_\mathrm{task}$
readout has finite reliability (its cross-model consistency, $+0.12$,
indicates the ceiling any per-unit correlate can reach), so true-scale
bounds are weaker; the cumulative-eviction analysis (\S\ref{sec:45}),
which aggregates over units, is the leg of the argument that does not
depend on per-unit reliability.

\section{Results}\label{sec:4}

\begin{figure}[t]
  \centering
  \includegraphics[width=\linewidth]{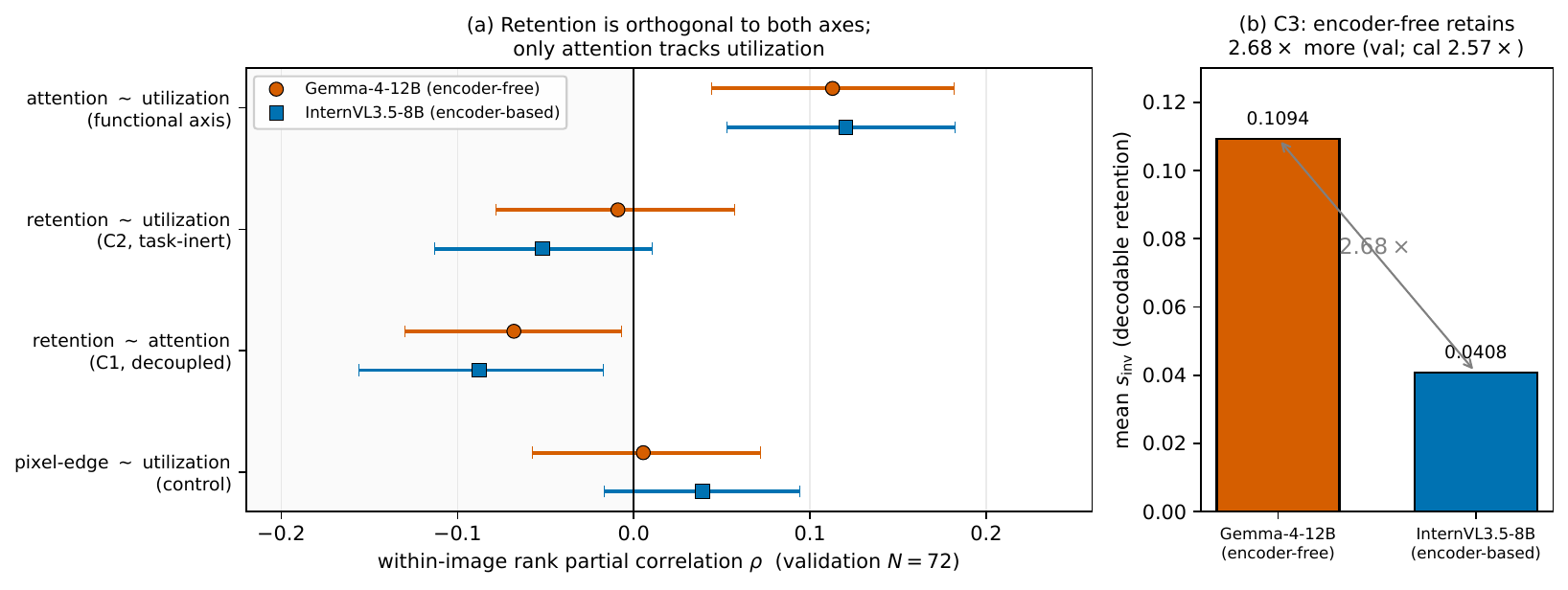}
  \caption{The dissociation results at a glance: (a) within-image partial-$\rho$ estimates for attention--utilization, C2, C1, and the pixel-edge control in both models; and (b) the C3 retention contrast (validation $2.68\times$, 95\% CI $[2.37, 3.06]$; calibration $2.57\times$).}
  \label{fig:forest}
\end{figure}

All correlations below are within-image rank partial correlations on the
held-out validation split (\(N=72\)), estimated by cluster bootstrap
over images (\(n=10{,}000\)); the sign of each hypothesized effect was
fixed on the disjoint calibration split (\(N=48\)). We report both
models (Gemma-4-12B, encoder-free; InternVL3.5-8B, encoder-based).
Figure~\ref{fig:forest} summarizes the four within-image
relationships and the architecture contrast at a glance;
Table~\ref{tab:correlations} gives the full intervals.

\begin{table}[t]
\caption{Within-image rank partial correlations, held-out validation ($N=72$ images, 1152 units), 95\% cluster-bootstrap CIs over images ($n=10{,}000$).}
\label{tab:correlations}
\centering
\small
\begin{tabularx}{\linewidth}{@{}>{\raggedright\arraybackslash}X >{\centering\arraybackslash}p{0.23\linewidth} >{\centering\arraybackslash}p{0.25\linewidth}@{}}
\toprule
Relationship & Gemma-4-12B (encoder-free) & InternVL3.5-8B (encoder-based) \\
\midrule
\textbf{C1} retention$\perp$attention $\rho(s_\mathrm{inv},s_\mathrm{attn})$ & $-0.068$ $[-0.130,-0.007]$ & $-0.088$ $[-0.156,-0.017]$ \\
\textbf{C2} retention$\perp$utilization $\rho(s_\mathrm{inv},\Delta_\mathrm{task}\mid \text{cov})$ & $-0.009$ $[-0.076,+0.057]$ & $-0.052$ $[-0.113,+0.010]$ \\
\textbf{attention$\sim$utilization} $\rho(s_\mathrm{attn},\Delta_\mathrm{task}\mid \text{pix,pos})$ & $+0.113$ $[+0.044,+0.182]$ & $+0.120$ $[+0.053,+0.183]$ \\
pixel-edge$\sim$utilization $\rho(s_\mathrm{pixel\text{-}edge},\Delta_\mathrm{task})$ & $+0.005$ $[-0.058,+0.072]$ & $+0.039$ $[-0.017,+0.094]$ \\
\bottomrule
\end{tabularx}
\par\smallskip
\begin{minipage}{\linewidth}
\footnotesize
C2 robustness (all agree on null): mixed random-intercept $\beta_\mathrm{inv}=+0.005$ / $-0.055$; per-image Wilcoxon $p=0.80$ / $0.41$. Reliability (split-half) $0.84$ / $0.89$. C2 power $\approx0.9$ at held-out $N=72$ vs SESOI $\rho=0.15$; upper CIs bound the effect below $\rho\approx0.06$ (Gemma) / $0.01$ (InternVL).
\end{minipage}
\end{table}

\subsection{Retention is decoupled from attention
(C1)}\label{sec:41}

Pixel-decodable retention and attention mass are, if anything, weakly
\emph{negatively} correlated within an image --- far below the
preregistered decoupling threshold of \(0.5\), and in the wrong
direction to serve as a proxy. The rank correlation
\(\rho(s_\mathrm{inv}, s_\mathrm{attn})\) is \(-0.068\) for Gemma
(one-sided 95\% bootstrap upper bound \(-0.017\); two-sided CI
\([-0.130, -0.007]\)) and \(-0.088\) for InternVL, holding
across both the calibration and validation splits. The KV positions that
retain the most decodable pixel content are not the positions the model
attends to: retention is not a proxy for attention (throughout,
``decoupled'' means far below the preregistered proxy threshold, not a
literally zero correlation).

\subsection{Retention is decoupled from causal utilization (C2,
well-powered
null)}\label{sec:42}

\paragraph{Result.} Retention is likewise uninformative about which units the model
\emph{causally uses}. On validation, the covariate-adjusted partial
correlation
\(\rho(s_\mathrm{inv}, \Delta_\mathrm{task} \mid s_\mathrm{attn}, s_\mathrm{pixel}, \text{pos})\)
is \(-0.009\) for Gemma (95\% cluster-bootstrap CI \([-0.076, +0.057]\),
tightly straddling zero) and \(-0.052\) for InternVL (CI
\([-0.113, +0.010]\), a weak negative whose interval still includes
zero). The direction that matters is the one hypothesized in advance and
fixed on calibration (\(\mathrm{sign_{cal}}=+1\): retention would
\emph{positively} predict utilization if reconstructable content were
preferentially used). Neither model shows it: both point estimates
are at or below zero, and the sign-oriented one-sided 95\% lower bound
is negative for both (Gemma \(-0.070\), InternVL \(-0.105\)). This is a
\emph{well-powered, bounded} null (\S\ref{sec:34}: power \(\approx 0.9\) at
\(N{=}72\) against \(\rho=0.15\); upper CIs exclude
\(\rho\gtrsim0.06\)/\(0.01\)), not an absence of evidence. Critically,
the null is not an artifact of an insensitive instrument: the
same within-image rank-partial estimation pipeline, on the same held-out split, detects a positive
attention--utilization association (\(\rho\approx+0.11\), CI excluding
zero; \S\ref{sec:43}). We therefore state a positive claim: within our setting,
pixel-decodable retention does not predict causal utilization (the
InternVL estimate leans weakly negative; Gemma's is essentially zero).

\paragraph{Robustness of the null.} Three families of checks make the
null robust rather than fragile. \emph{Design-internal checks.} The sign
fixed on calibration was weakly \emph{positive} for Gemma (\(+0.052\);
the InternVL calibration estimate was \(-0.005\), essentially zero and
uninformative about direction) yet did not replicate on validation --- consistent with an effect
smaller than the between-split variation. The retention instrument is reliable
(split-half \(0.84\)/\(0.89\) across decoder seeds; \(0.82\)/\(0.88\)
for the exact within-image-ranked, covariate-residualized quantity
entering the estimator), so the null cannot be explained away as
measurement or residualization noise, and our conservative
disattenuation (\S\ref{sec:34}) only pushes the estimate further from
--- not toward --- zero. \emph{Estimator robustness.} A mixed-effects
model with per-image random intercepts yields retention coefficients of
\(+0.005\)/\(-0.055\), and a per-image Wilcoxon test fails to reject
zero for both models (\(p=0.80\)/\(0.41\)): three estimators, one
conclusion. \emph{Measurement-choice robustness.} The null holds across
decoder depth --- recomputing \(s_\mathrm{inv}\) for Gemma from
layer-0/6/12 KV, each with its own decoder, yields \(\rho = -0.009\),
\(-0.020\), \(-0.039\) (all CIs include zero; never the hypothesized
positive coupling), even as decodable retention itself weakens with
depth (mean \(s_\mathrm{inv}\) \(0.108\to0.088\to0.066\), two-seed
ensemble; the architecture contrast in Figure~\ref{fig:forest}b uses
the seed-wise estimate, \(0.109\)), and C1
likewise holds at all three depths --- so the task-inertness is not a
primary-layer artifact (Gemma; the encoder-based decoder was trained
only at layer 0). Nor is it manufactured by deconfounding: a progressive
covariate ladder shows the \emph{marginal} correlation is already
\(\approx 0\) for Gemma (\(-0.004\) \([-0.066,+0.059]\)) and weakly
\emph{negative} for InternVL (\(-0.068\) \([-0.131,-0.010]\)) --- the
covariates absorb a negative texture confound, not a positive signal.
Finally, \(\Delta_\mathrm{task}\) itself carries genuine cross-model
structure: the two models' within-image utilization maps correlate at
\(+0.12\) \([+0.05,+0.20]\) on identical images and questions, bounding
the readout's reliability away from zero.

\subsection{Attention weakly but genuinely tracks
utilization}\label{sec:43}

Crucially, the utilization axis is \emph{not} inert to every proxy:
attention weakly predicts it. The partial correlation
\(\rho(s_\mathrm{attn}, \Delta_\mathrm{task} \mid s_\mathrm{pixel}, \text{pos})\)
is \(+0.113\) for Gemma (CI \([+0.044, +0.182]\)) and \(+0.120\) for
InternVL (CI \([+0.053, +0.183]\)), both positive with held-out
intervals that exclude zero. By contrast, pixel-energy covariates do not
predict utilization
(\(\rho(s_\mathrm{pixel\text{-}edge}, \Delta_\mathrm{task}) = +0.005 / +0.039\),
CIs contain zero), ruling out ``textured patches matter'' as the
explanation. This result does double duty. It confirms the
\(\Delta_\mathrm{task}\) instrument carries real signal (something ---
attention --- predicts it), which in turn hardens the C2 null
(\S\ref{sec:42}). And it
establishes an asymmetry: of the axes we measure, attention is the
\emph{only} one with a statistically detectable relationship to
utilization, and pixel-decodable retention is orthogonal to it. At \(\rho\approx0.11\) attention
explains only about one percent of the within-image variance in
\(\Delta_\mathrm{task}\): it is the better proxy, not a good one. We
use ``functional axis'' as shorthand for this signal --- the only
utilization-tracking signal we detect; the substantive claim is
comparative: retention tracks utilization even less,
indistinguishably from zero. In that comparative sense pixel-decodable
retention behaves as an \textbf{informational axis} of the cache
(content carried and extractable) that is orthogonal to the weak
functional signal (content the computation acts on). One structural
asymmetry deserves note when reading this comparison: retention is by
construction a query-agnostic readout (image-only prefill), while
attention and utilization are both measured under the
question-conditioned pass. An eviction score must be computable before
the answer in any case, so the comparison reflects deployment reality,
but ``the only proxy that tracks utilization'' is also the only
query-conditioned one (Appendix~\ref{app:crossq} examines how much
query-conditioning attention actually uses).

\subsection{Retention differs across the architecture pair
(C3)}\label{sec:44}

How much task-inert pixel content a cache retains differs sharply
between our two models. At the primary layer, the encoder-free Gemma
retains \(2.68\times\) more decodable pixel information per unit than
the encoder-based InternVL (cluster-bootstrap 95\% CI
\([2.37, 3.06]\); calibration \(2.57\times\), consistent
across splits and decoder seeds). A natural mechanistic reading is that,
without a ViT bottleneck to re-encode the image first, pixel content
enters the language model directly and persists there in decodable form
--- even though, per \S\ref{sec:42}, that persistence buys no additional causal
utility. This mechanistic reading remains a model-pair hypothesis: with one model per
architecture class, the \(2.68\times\) is confounded with every other
difference between Gemma-4-12B and InternVL3.5-8B (parameter count,
training data, tokenizer, visual token budget), so we report it as a
\emph{model-pair} contrast and defer an architecture-\emph{class} claim
to the multi-model replication discussed in \S\ref{sec:5}.

\subsection{Compression implication: retention is not a pruning
signal}\label{sec:45}

\begin{figure}[t]
  \centering
  \includegraphics[width=\linewidth]{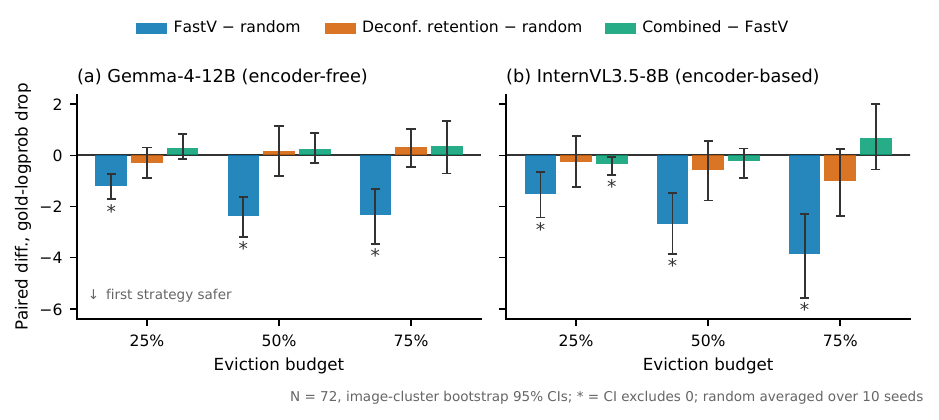}
  \caption{Eviction contrasts on the super-patch grid (validation,
  $N{=}72$): paired differences in gold-answer logprob drop at each
  budget (negative = first strategy safer; error bars = image-cluster
  bootstrap 95\% CIs; $^*$ = CI excludes zero, as in
  Tables~\ref{tab:compression} and~\ref{tab:token}; random averaged
  over 10 seeds). FastV $-$ random excludes zero at every budget
  (attention helps); deconfounded retention $-$ random contains zero
  at every budget (no super-patch compression signal); combined $-$
  FastV shows no consistent gain, with the one exception of InternVL
  at the 25\% budget. Full grid: Table~\ref{tab:compression};
  token-granularity contrasts: Table~\ref{tab:token}.}
  \label{fig:pareto}
\end{figure}

The dissociations above predict a concrete engineering consequence.
Reconstruction-based importance admits two rival readings, and we test
both. Under the \emph{removable-redundancy} reading, a unit whose pixel
content is highly reconstructable is redundantly stored and therefore
safe to drop: evict the \emph{highest}-retention super-patches. Under
the \emph{high-value} reading, high retention marks content worth
keeping: evict the \emph{lowest}-retention super-patches instead. If
retention is task-inert, \emph{neither} rule should beat evicting at
random. We test it directly with a
controlled KV-eviction Pareto on validation (\(N=72\)). For each
strategy we evict \(k\in\{4,8,12\}\) of the 16 super-patches' visual KV
in a single fresh forward pass and measure the teacher-forced
gold-answer log-probability drop (lower is safer) and greedy-decoded VQA
accuracy; random is averaged over 10 seeds; all differences use an
image-level cluster bootstrap (\(n=2{,}000\)). Strategies: attention-guided
\textbf{FastV} (evict lowest-attention), \textbf{raw retention} (evict
highest \(s_\mathrm{inv}\)), \textbf{deconfounded retention} (evict
highest within-image-residualized \(s_\mathrm{inv}\)), \textbf{combined}
(the sum of the within-image retention rank and the reversed attention
rank, so that patches that are simultaneously high-retention and
low-attention are evicted first), and \textbf{random}.
Table~\ref{tab:compression} gives the full grid; the paired contrasts
tell the story, and Figure~\ref{fig:pareto} shows them.

\begin{table}[!ht]
\caption{Compression Pareto --- KV eviction on validation ($N=72$). Each
cell: mean gold-logprob drop (lower = safer) / VQA accuracy, per strategy
$\times$ budget $k$; random averaged over 10 seeds; paired differences via
image-cluster bootstrap ($n{=}2{,}000$). Bold marks the attention baseline (FastV).}
\label{tab:compression}
\centering
\small
\begin{tabular*}{\linewidth}{@{\extracolsep{\fill}}lccc@{}}
\toprule
\multicolumn{4}{c}{\emph{Gemma-4-12B (encoder-free; baseline VQA 0.66)}} \\
\midrule
strategy & $k{=}4$ & $k{=}8$ & $k{=}12$ \\
\midrule
\textbf{FastV} (attention) & \textbf{+0.25 / 0.60} & \textbf{+0.68 / 0.56} & \textbf{+2.84 / 0.36} \\
Raw retention & +0.77 / 0.57 & +2.40 / 0.44 & +4.56 / 0.22 \\
Deconfounded retention & +1.16 / 0.54 & +3.20 / 0.40 & +5.49 / 0.22 \\
Combined & +0.53 / 0.54 & +0.93 / 0.51 & +3.20 / 0.34 \\
Random & +1.46 / 0.45 & +3.04 / 0.33 & +5.17 / 0.19 \\
\midrule
\multicolumn{4}{c}{\emph{InternVL3.5-8B (encoder-based; baseline VQA 0.73)}} \\
\midrule
strategy & $k{=}4$ & $k{=}8$ & $k{=}12$ \\
\midrule
\textbf{FastV} (attention) & \textbf{+0.30 / 0.67} & \textbf{+0.83 / 0.61} & \textbf{+2.49 / 0.56} \\
Raw retention & +0.63 / 0.62 & +2.33 / 0.50 & +5.46 / 0.31 \\
Deconfounded retention & +1.54 / 0.57 & +2.95 / 0.44 & +5.35 / 0.28 \\
Combined & $-0.04$ / 0.69 & +0.60 / 0.61 & +3.18 / 0.50 \\
Random & +1.79 / 0.57 & +3.53 / 0.46 & +6.35 / 0.29 \\
\bottomrule
\end{tabular*}
\end{table}

\FloatBarrier

Paired contrasts on the super-patch grid (mean diff [95\% CI]; $*$ = CI
excludes 0; differences computed from unrounded per-image values):
\begin{itemize}
\item \textbf{Deconfounded retention $-$ random} (main; expect $\approx0$): Gemma $-0.30$/$+0.16$/$+0.32$, InternVL $-0.25$/$-0.57$/$-1.01$ --- \textbf{all include 0} (no super-patch compression signal).
\item \textbf{FastV $-$ random}: Gemma $-1.21^*$/$-2.36^*$/$-2.33^*$, InternVL $-1.49^*$/$-2.70^*$/$-3.86^*$ --- \textbf{all exclude 0} (attention helps).
\item \textbf{Combined $-$ FastV}: Gemma $+0.28$/$+0.25$/$+0.35$, InternVL $-0.34^*$/$-0.22$/$+0.69$ --- no consistent gain from adding retention. The one significant exception is InternVL at $k{=}4$, where the combined rule beats FastV on both metrics ($-0.04$ vs.\ $+0.30$ cost; $0.69$ vs.\ $0.67$ accuracy) --- retention can complement attention at small budgets on one model, but the gain does not persist at larger budgets or on Gemma.
\end{itemize}

\textbf{Deconfounded retention ranks eviction no better than random at
super-patch granularity --- on both models.} The paired difference
deconfounded retention $-$ random has a 95\% CI containing zero at
every budget for both VLMs; at the larger Gemma budgets
the point estimate is even mildly \emph{positive} (deconfounded
retention slightly worse than random). By contrast, \emph{raw} retention
does appear to beat random at the smallest budget
(raw retention $-$ random excludes zero: Gemma \(-0.69\),
InternVL \(-1.16\)), but that apparent advantage evaporates the
moment retention is residualized within-image against texture,
attention, and position. This is the compression-side confirmation, on a
real cumulative-eviction forward pass, of the \S\ref{sec:42} null: retention's
surface-level compression value has no detectable unique advantage once
within-image confounds are removed. The paired super-patch intervals also
\emph{bound} what remains undetected: the maximum safety advantage of
deconfounded retention over random compatible with the 95\% CIs
($0.88/0.77/0.45$ gold-logprob for Gemma at $k{=}4/8/12$;
$1.28/1.71/2.40$ for InternVL) is smaller at every budget than the
advantage FastV actually realizes ($1.21$--$2.36$ and $1.49$--$3.86$
respectively) --- so, at this granularity, ``no better than random'' is
not merely a failure to reject, but an interval statement placing
retention's best case below attention's observed case. One bookkeeping caveat for Gemma: because its
token-to-super-patch blocks are non-uniform (\S\ref{sec:33}), the
\emph{realized} evicted-token fraction at a nominal budget varies
slightly by strategy (per-strategy means $22.8\%$--$25.2\%$ at
$k{=}4$; FastV in fact evicts marginally \emph{fewer} tokens than the
retention rules), a $\le 2.4$ percentage-point wobble that slightly
favors the retention rules' safety and therefore cannot explain their
failure to beat random. The secondary greedy-VQA-accuracy
metric tells the same story with one wrinkle: paired
deconfounded retention $-$ random accuracy differences include zero
at every budget for InternVL and at Gemma's two larger budgets, with a
single marginal exception at Gemma's smallest budget (\(+0.09\), CI
\([+0.02,+0.16]\)) that would not survive Bonferroni correction across
the twelve strategy\(\times\)budget contrasts and in any case points the
wrong way for retention-as-importance: its sign favors \emph{evicting}
high-retention patches; on the super-patch grid the removable-redundancy
reading that would support is rejected on the primary metric at every
budget (at token granularity a weak effect in that direction does
emerge; see below). The complementary policy of treating retention as a
\emph{keep} signal and evicting the \emph{lowest}-retention units
instead fares no better: on an additive per-unit proxy (summing
single-unit \(\Delta_\mathrm{task}\) over the evicted set) on the same
validation images, evicting the \(k\) lowest-retention super-patches is
indistinguishable from random for Gemma and significantly \emph{worse}
than random for InternVL (e.g., \(+1.78\) at \(k{=}8\), CI excluding
zero; raw and deconfounded scores behave alike).
Neither direction of the retention ranking carries eviction value, so
the null is not an artifact of having tested the ``wrong end'' of the
score.

\textbf{Token granularity.} All results above operate on the
16-super-patch grid. Re-running the cumulative-eviction protocol at
native token granularity (252--275 visual tokens per image for Gemma,
256 for InternVL; budgets of 25/50/75\% of each image's actual token
count, so at a given budget every strategy deletes exactly the same
number of tokens) sharpens the comparative picture and triggers one
preregistered narrowing (Table~\ref{tab:token}). The attention positive
control holds: FastV $-$ random excludes zero at every budget on both
models. Deconfounded retention, in contrast with the super-patch null,
acquires a weak \emph{inverse}-importance signal: evicting the
highest-retention tokens is safer than random at the 50\% and 75\%
budgets on both models (Gemma $-1.29$ $[-2.02,-0.55]$ and $-0.97$
$[-1.79,-0.18]$; InternVL $-1.13$ $[-2.34,-0.10]$ and $-1.31$
$[-2.42,-0.30]$ gold-logprob), while the 25\% intervals include zero and the
effect is not monotone in budget (Gemma's 50\% effect exceeds its
75\%); a per-seed check confirms no single random seed drives it, and
raw retention shows the same directional pattern (its 75\% intervals
marginally include zero). The
preregistered narrowing therefore applies: ``no better than random'' is
a super-patch-level statement, and at token granularity high
pixel-decodable retention weakly marks \emph{safe-to-evict} content ---
the direction opposite to what retention-as-importance would need.
Unchanged, and now strengthened from an interval bound to a direct
paired comparison, is the ordering that matters for practice: FastV $-$
deconfounded retention excludes zero at every budget on both models, so
attention dominates retention wherever the two disagree. The secondary
VQA metric agrees in direction with the retention\(-\)random inversion
where it is significant (Gemma's two smaller budgets; no InternVL
budget reaches significance). The spatial
counterpart of this inversion --- retention anti-localizing
OCR-grounded answer regions --- is validated independently in
Appendix~\ref{app:criterion}.

\begin{table}[!ht]
\caption{Token-granularity eviction, paired gold-logprob contrasts
(mean difference; negative = first strategy safer; image-cluster
bootstrap, $n{=}2{,}000$; random averaged over 5 seeds within image;
budgets are 25/50/75\% of each image's actual visual-token count and
identical across strategies). $^*$ = 95\% CI excludes 0; key intervals
in the text.}
\label{tab:token}
\centering
\small
\begin{tabular*}{\linewidth}{@{\extracolsep{\fill}}llccc@{}}
\toprule
Model & Contrast & 25\% & 50\% & 75\% \\
\midrule
Gemma & Deconf.\ retention $-$ random & $-0.17$ & $-1.29^*$ & $-0.97^*$ \\
Gemma & FastV $-$ random & $-1.53^*$ & $-3.61^*$ & $-4.52^*$ \\
Gemma & FastV $-$ deconf.\ retention & $-1.36^*$ & $-2.33^*$ & $-3.56^*$ \\
\midrule
InternVL & Deconf.\ retention $-$ random & $-0.25$ & $-1.13^*$ & $-1.31^*$ \\
InternVL & FastV $-$ random & $-1.08^*$ & $-2.25^*$ & $-4.01^*$ \\
InternVL & FastV $-$ deconf.\ retention & $-0.83^*$ & $-1.12^*$ & $-2.70^*$ \\
\bottomrule
\end{tabular*}
\end{table}

\textbf{Attention is the weak-but-real signal; retention adds no
consistent gain.} Attention-guided FastV is robustly the safest single-proxy strategy
--- FastV $-$ random excludes zero at every budget for both
models, and it also retains the most VQA accuracy (Table~\ref{tab:compression}). Adding retention to attention buys no consistent improvement:
combined $-$ FastV contains zero at nearly every budget (and
trends positive for Gemma). On the primary causal-cost metric the
super-patch ordering is unambiguous --- attention \(>\) random
\(\approx\) deconfounded retention --- and at token granularity
attention dominates deconfounded retention at every budget
(Table~\ref{tab:token}). It mirrors the correlational result exactly:
attention weakly tracks utilization (\S\ref{sec:43}) and helps
compression; pixel-decodable retention positively tracks neither ---
at its strongest it weakly marks what is \emph{safe to discard}. What
the cache reconstructably retains is not what is worth keeping.

\section{Discussion}\label{sec:5}

\textbf{Why is decodable content retained but not used?} The
dissociation is less paradoxical than it first appears once retention
and utilization are seen as answering different questions about the same
KV vector. Our inversion decoder asks whether pixel content is
\emph{recoverable} from a unit's KV by a learned (non-linear) decoder;
causal ablation asks whether the model's forward computation
\emph{routes the answer through} that unit. A representation can be rich
in recoverable content yet not be the locus the computation reads from.
The architecture result (\S\ref{sec:44}) suggests a mechanistic reading. In an
encoder-free VLM, pixels enter the language model directly and their
content remains decodable at the layer we probe, with no ViT bottleneck
to re-encode them first. But being decodable at a spatial unit is not
the same as being \emph{used} per spatial unit: the information relevant
to the answer is mixed across positions, so removing any single
super-patch's KV rarely severs a load-bearing path (\S\ref{sec:42}), even as the
raw content sits there to be decoded. Attention, meanwhile, is trained
toward task relevance rather than information content, which is why it
--- the only proxy we measure that does --- weakly tracks utilization
(\S\ref{sec:43}). Retention measures what the cache \emph{contains}; attention
approximates what the computation \emph{prefers}; and, among the
proxies we measure, only the latter
bears any relation to what the computation \emph{needs}.

\textbf{From probing principle to cache-compression evidence.} The
established decodable-\(\neq\)-causal lens (\S\ref{sec:2}) becomes
operational here in three ways. First, the relevant hypothesis is not folklore but
\emph{practice}: KV- and token-compression methods do use
reconstruction- and information-content signals, and one cannot know a
priori whether pixel-decodable retention is among the signals that work
--- we show, causally, that it is not. Second, the effect is quantified,
not asserted: a preregistered, well-powered, bounded null (not ``we did
not find an effect'') with a working positive control on the same
estimator, which distinguishes genuine inertness from an insensitive
measurement. Third, the null is not global --- attention \emph{does}
track utilization --- so the finding is a specific claim about
\emph{which} axis is inert, not a blanket ``nothing correlates.'' The
contribution is a map and a causal instrument for drawing it, of which
this dissociation provides one concrete entry.

\textbf{Implications for KV-cache compression.} The engineering reading
of these results is a caution against reconstruction- or
information-content-based importance scores for visual KV eviction. In
our setting, deconfounded pixel-decodable retention ranks eviction no
better than random at super-patch granularity (and helps at token
granularity only through its weak inverse signal), whereas attention
magnitude --- the weak-but-real proxy --- does at every granularity. Neither signal is strong (attention itself only reaches
\(\rho\approx0.11\) with utilization), which suggests that, in our
setting, single-proxy visual-cache pruning has a low ceiling and that
causal or task-aware signals may be needed to push past it. A direct
test reinforces the ceiling reading: a ridge combination of thirteen
readily available per-unit signals (multi-layer attention variants, OCR
density, position, pixel statistics), supervised on the calibration
split by the causal \(\Delta_\mathrm{task}\) readout itself, fails to
outperform plain attention on held-out validation (\(\rho = 0.07/0.06\)
vs.~\(0.09/0.13\); within-image rank correlations without covariate
partialling, hence differing slightly from Table~\ref{tab:correlations})
--- the extractable per-unit importance signal in
these features is essentially exhausted by attention alone. Importantly,
this is not a claim against \emph{feature-space} reconstruction methods,
which reconstruct a different quantity and report real gains; our
negative result is specific to pixel-level decodability used as an
eviction score.

\textbf{What attention selects is nearly question-invariant.} A
post-hoc analysis (Appendix~\ref{app:crossq}) adds a structural
dimension to the low single-proxy ceiling. On the validation images
that carry two official questions, the retained sets an attention-based
policy would keep are nearly identical across questions (token-level
Jaccard $0.87$--$0.95$ for Gemma and $0.74$--$0.94$ for InternVL across
budgets, against random-set expectations of $0.14$--$0.60$), while the
per-unit utilization readouts under the two questions correlate only
weakly ($\rho \approx 0.09$, interval including zero --- bounded by
per-unit reliability, so we do not conclude query-specificity;
Appendix~\ref{app:crossq}). Read together with \S\ref{sec:45} and
Appendix~\ref{app:criterion}, the picture is coherent: the one weak
functional proxy operates as question-independent salience rather than
question-conditioned routing, and the retention axis relates to task
structure only with inverted sign --- spatially (it anti-localizes
answer regions) and behaviorally (evicting its highest-scoring tokens
is safer than random at the larger budgets). Both observations mark
out the same boundary: per-unit, query-agnostic signals --- whether
informational or attentional --- appear to face a structural ceiling
in our setting, which is precisely where query-conditioned or causal
signals would have to enter.

\textbf{Limitations.} Our claims are deliberately scoped to the setting
we measured: two VLMs (one encoder-free, one encoder-based), the TextVQA
task, super-patch granularity, and a primary decoder layer. Several of
these are load-bearing. The utilization readout
is a single-unit ablation; importance that is redundantly distributed
across units is not captured by the per-unit \(\Delta_\mathrm{task}\),
and for the encoder-based model the ViT's early mixing makes it a
conservative lower bound (\S\ref{sec:32}) --- the cumulative-eviction analysis of
\S\ref{sec:45} and the essentially-zero encoder-free estimate partly offset both concerns.
TextVQA localizes answers to specific regions, which is what makes
single-unit ablation informative --- tasks with more diffuse visual
dependence may show a different retention/utilization relationship. The architecture contrast rests on one model
per class, so the \(2.68\times\) is a model-pair effect confounded with
model scale and training; extending to additional encoder-free and
encoder-based VLMs is what would license an architecture-\emph{class}
claim. The decoder-layer choice, by contrast, we tested directly:
re-running the Gemma analysis at layers 0, 6, and 12 leaves the
retention--utilization null intact at every depth (\S\ref{sec:42}), so the
task-inertness is not a primary-layer artifact; extending this depth
sweep to the encoder-based model (whose secondary-layer decoders we did
not train) remains future work. Finally, the cross-question invariance
of attention-selected sets (\S\ref{sec:5}, Appendix~\ref{app:crossq})
is itself a scoped, post-hoc observation on the official TextVQA
question pairs, not a preregistered result.

\subsubsection*{Broader impact statement}
This work evaluates measurement validity of importance signals for
visual KV-cache compression. Its practical effect is to steer
deployed-efficiency work away from a signal that does not withstand a
causal test; we do not foresee direct negative societal impacts beyond
those of VLM deployment generally.

\subsubsection*{Reproducibility statement}
All quantities reported are computed from the official TextVQA
validation set with preregistered splits and seeds; code, the per-unit
measurement tables, the eviction harness, the frozen preregistration
protocol (SESOI, thresholds, split procedure, sign-freezing rule), and
the exact model checkpoints and revisions will be released.

\bibliography{refs}
\bibliographystyle{tmlr}

\appendix
\section{Post-hoc exploratory: cross-question stability of the attention axis}\label{app:crossq}

\emph{The analyses in this appendix were conducted after the
preregistered study concluded, on the subset of validation images that
carry two official TextVQA questions ($N=34$); they are exploratory and
were not preregistered.}

A natural question raised by \S\ref{sec:43} and \S\ref{sec:45} is
whether the weak-but-real attention signal is \emph{query-conditioned}
in what it selects. We measured this directly: for each of the 34
validation images with exactly two official questions, we computed the
attention score under each question (identical estimand to \S\ref{sec:sattn}) and
compared the retained sets an attention-based eviction policy would
keep.

\paragraph{Attention retained sets are nearly invariant across
questions.} At token granularity --- the granularity at which eviction
methods operate --- the Jaccard overlap between the top-$b$ sets under
the two questions is $0.87/0.91/0.95$ for Gemma and $0.74/0.86/0.94$
for InternVL at $b=25\%/50\%/75\%$ budgets (cluster-bootstrap CIs
within $\pm 0.04$; random-set expectation $0.14/0.33/0.60$).
Super-patch granularity gives the same picture, so the invariance is
not an aggregation artifact. Nor is it an attention-\emph{sink}
artifact: stripping the most sink-like positions (up to 20\% of
positions, ranked by image-independent mean attention rank) leaves the
overlap essentially unchanged (e.g., InternVL $0.745 \to 0.728$ at the
25\% budget), and only ${\sim}20\%$ of the cross-question intersection
consists of sink positions --- the invariance is carried by
image-driven, not position-driven, attention. Qualitatively, a
query-independent component of visual attention has been reported
before \citep{attnsink}; the observation here is quantitative and
selection-level: \emph{the sets an eviction policy retains barely
change with the question} (in our setting: TextVQA official pairs,
two questions per image).

\paragraph{What this does and does not imply.} The modal gold answers
of the two questions differ for 88\% of these images, and the two
questions' causal-utilization maps correlate only weakly ($\rho =
0.086$ $[-0.028, +0.203]$). We explicitly do \emph{not} conclude from
this that utilization is query-specific: the per-unit
$\Delta_\mathrm{task}$ readout's own reliability bounds what any such
cross-question correlation can reach --- cross-\emph{model}
same-question consistency ($+0.117$ $[+0.004, +0.222]$ on this 34-image
subset; $+0.12$ $[+0.05,+0.20]$ on the full validation set,
\S\ref{sec:42}) is of the same
magnitude as cross-\emph{question} same-model consistency, so
unit-level measurement noise cannot be excluded as the driver of both.
One diagnostic did resolve cleanly: the same-question control on the
second questions ($\rho(s_\mathrm{attn}, \Delta_\mathrm{task}) =
0.029$, anomalously below the ${\approx}0.11$ of \S\ref{sec:43}) is
driven by low-annotator-agreement questions --- in the high-agreement
stratum the control recovers to $0.101$, matching the first-question
baseline ($0.109$) on the same images. The instrument behaves normally
when the gold target is clean.

\paragraph{Reading.} The selection-level stability of attention across
queries is consistent with the picture of \S\ref{sec:4}: attention-based
eviction (FastV) is robust (\S\ref{sec:45}) yet couples to causal
utilization only weakly (\S\ref{sec:43}) --- what it selects is a
largely query-independent notion of visual salience. We report this as
a scoped, post-hoc observation on official question pairs of a
text-reading task; whether it persists under deliberately
region-divergent queries (e.g., multi-page document QA) is an open
question we do not answer here.

\section{Instrument details: the pixel-inversion decoder and $s_\mathrm{inv}$}\label{app:instrument}

\paragraph{Decoder.} For each probed layer $\ell$, the inversion
decoder maps the layer-$\ell$ visual KV of an image-only prefill back
to pixel space. Each 4096-dimensional KV token is projected linearly to
a 48-channel spatial cell on the token grid; a convolutional stack
(Conv$\to$GELU, ConvTranspose upsampling $\times 2$ twice, interleaved
$3{\times}3$ convolutions) decodes the grid to a 3-channel image at 16
pixels per token cell ($\approx$3.37M parameters per layer). Decoders
are trained with a pixel MSE objective (Adam, learning rate
$3{\times}10^{-4}$, 40 epochs, batch size 4) on 200 images from the
official TextVQA \emph{training} split (50 held out for decoder
validation), disjoint by official split from every image used in
calibration or validation. Two decoders are trained per layer under
independent seeds; the analysis uses their within-image z-scored
ensemble, and the seed pair provides the split-half reliability used for
disattenuation (\S\ref{sec:34}).

\paragraph{Reconstruction score and $s_\mathrm{inv}$.} Reconstruction
quality is measured by \emph{gradient correlation}: the Pearson
correlation between the concatenated horizontal and vertical spatial
gradients of the decoded image and those of the ground-truth image,
computed over the pixel region of a super-patch. For each super-patch
$u$, the retention score is the shuffle-adjusted excess
$s_\mathrm{inv}(u) = g_u(\mathrm{KV}) -
\overline{g_u(\mathrm{KV}_\pi)}$, where $g_u$ is the regional gradient
correlation with the KV intact and the second term averages three
within-image random permutations $\pi$ of KV positions. The shuffle
baseline removes what any KV content would afford at that output
location, so $s_\mathrm{inv}$ isolates position-specific decodable
content; it is a dimensionless correlation difference, and the C3
contrast (\S\ref{sec:44}) compares its per-unit means across models
(ratio $2.68\times$, cluster-bootstrap 95\% CI $[2.37, 3.06]$).

\section{Criterion validation: localizing OCR-grounded answer
regions}\label{app:criterion}

The ablation readout $\Delta_\mathrm{task}$ is our measure of causal
utilization; a natural question is whether it has spatial validity of
its own --- whether high-$\Delta_\mathrm{task}$ units coincide with
where the task-relevant content actually sits. We validate this against
an answer-localization criterion defined independently of any
model-derived score: official Rosetta OCR boxes for the TextVQA
validation images, matched to the answer string.

\textbf{Setup.} The preregistered protocol matches the normalized gold
answer against OCR tokens with a three-tier rule (exact; multi-word
subsequence; edit distance $\le 2$) and maps matched boxes to
super-patches under a center-point criterion (IoU $\ge 0.1$ as
sensitivity). The implementation additionally admitted the ten
normalized ground-truth answer variants as match candidates --- a
deviation from the preregistered gold-only wording that we report as
the expanded surface, keeping the preregistration-faithful gold-only
restriction as the primary reading; the two agree, and gold-only is
stronger. Gold-only localizes the answer in $N{=}52$ of 72 held-out
images (expanded surface $N{=}59$; positives per image 1:44, 2:11,
3:3, 4:1 on the expanded surface); unmatched images are spelling
variants and reasoning-type answers, and they do not differ detectably
from matched images at baseline (gold log-probability and VQA-accuracy
differences contain zero in the three model$\times$metric cells with
stored generative baselines; the Gemma VQA cell has none in the
archived artifacts). For each image we rank the 16 super-patches by
each score and compute answer-region AUC, aggregated with image-cluster
bootstrap.

\textbf{$\Delta_\mathrm{task}$ localizes the answer region on both
models.} Gold-only AUC $0.791$ $[0.703, 0.878]$ (Gemma) and $0.766$
$[0.670, 0.849]$ (InternVL); expanded surface $0.723$ $[0.626, 0.816]$
and $0.714$ $[0.623, 0.808]$; all four exclude $0.5$. The per-image
paired contrast (answer vs.\ non-answer mean $\Delta_\mathrm{task}$,
expanded surface) is $3.23$ $[2.06, 4.53]$ and $4.16$ $[2.09, 6.71]$.
The ablation readout is thus anchored to an external criterion, not
merely internally consistent. Attention localizes as well
($0.731$/$0.800$, expanded surface), consistent with \S\ref{sec:43}.
The pixel-edge control also localizes answer regions ($0.707$, both
models) --- text sits where texture is dense --- which is precisely why
the design residualizes against texture (\S\ref{sec:34}); localizing
the answer is not the same as predicting $\Delta_\mathrm{task}$
(\S\ref{sec:43}).

\textbf{The retention score's answer-location association is
inverted.} We preregistered any departure of the retention AUC from
$0.5$ as an alarm on the C2 measurement, to be reported regardless of
direction. The alarm fired, in the direction \emph{away} from answer
regions: AUC $0.393$ $[0.322, 0.463]$ (Gemma) and $0.415$
$[0.343, 0.485]$ (InternVL) on the expanded surface ($0.408$/$0.371$
gold-only). Two things follow. First, this is real, sign-invertible
discriminative information: retention is not information-free about
answer location. Second, it does not overturn the preregistered C2
result, which is a null of the \emph{continuous}
retention--utilization partial correlation; what the inverted AUC adds
is that where retention relates to task-relevant location at all, the
association points opposite to the direction any
retention-as-importance proxy would need --- consistent with the weak
negative retention--attention correlation (C1) and with the
token-granularity eviction result (\S\ref{sec:45}).

\textbf{Robustness.} Dropping the edit-distance tier gives
$0.734$/$0.713$ on the expanded surface ($N{=}44$) and $0.846$/$0.759$
gold-only ($N{=}32$); the IoU criterion agrees in direction ($N{=}26$:
$0.770$/$0.633$, the InternVL interval widening at this reduced $N$ to
include $0.5$). The gold-only AUC cells were recomputed with an
independent rank-sum implementation with exact agreement (8 summary and
416 per-image cells).

\end{document}